\documentclass[11pt]{article}

\usepackage[margin=1in]{geometry}
\usepackage[T1]{fontenc}
\usepackage[utf8]{inputenc}
\usepackage{amsmath,amssymb,amsthm}
\usepackage{enumitem}
\usepackage{booktabs}
\usepackage[hidelinks]{hyperref}

\newtheorem{theorem}{Theorem}[section]
\newtheorem{proposition}[theorem]{Proposition}
\newtheorem{lemma}[theorem]{Lemma}
\theoremstyle{definition}
\newtheorem{definition}[theorem]{Definition}
\newtheorem{question}{Question}[section]
\theoremstyle{remark}
\newtheorem{remark}[theorem]{Remark}

\newcommand{\K}{\mathbb{K}}
\newcommand{\R}{\mathbb{R}}
\newcommand{\C}{\mathbb{C}}
\newcommand{\fbar}{\bar f}
\newcommand{\Mult}{\mathsf{M}}

\title{Riordan Arrays as Structured Operators for Artificial Intelligence\\[4pt]
\large A Preliminary Research Discussion}

\author{
Dennis Davenport\\
Department of Mathematics\\
Howard University\\
Washington, DC, USA\\
\texttt{dennis.davenport@howard.edu}
\and
Dewayne A. Dixon\\
Department of Mathematical Sciences\\
Clark Atlanta University\\
Atlanta, GA, USA\\
\texttt{ddixon@cau.edu}
\and
Moussa Doumbia\\
Department of Mathematics\\
Howard University\\
Washington, DC, USA\\
\texttt{mdoumbia@howard.edu}
}

\date{}

\begin{document}
\maketitle

\begin{abstract}
Riordan arrays have traditionally been studied as infinite lower-triangular matrices that
encode generating-function transformations and combinatorial relationships. This note proposes
a complementary direction: using Riordan arrays as structured operators in artificial intelligence
and data analysis. For a data sequence with generating function $X(z)$, a Riordan array
$(g(z), f(z))$ produces the transformation
\[
X(z) \longmapsto g(z)\,X(f(z)).
\]
This gives a parameterized family of causal linear operators that combine convolution with
formal substitution, so that the kernel applied to each input coefficient depends on its index
in an algebraically constrained way. The resulting structure may support parameter-efficient
neural-network layers, variable-length sequence models, interpretable recurrent systems,
invertible architectures, and models for hierarchical or recursively generated data.

We define the Riordan-theoretic objects that the proposal relies on, explain what the
infinite-dimensional viewpoint does and does not provide, and record four elementary
observations: the pair $(g,f)$ is identifiable from two input--output pairs; sums of Riordan
operators leave the Riordan class; truncated Riordan layers have closed-form determinants, and
pseudo-involutory layers have condition number equal to the square of their norm; and a Riordan
layer can be applied without forming its matrix by one power-series composition and one
multiplication. We then outline several possible architectures, identify theoretical and
numerical challenges, and propose an initial experimental program. The aim is not to claim that
Riordan arrays are appropriate for all data, but to identify settings in which their algebraic
structure supplies a useful inductive bias.
\end{abstract}

\section{Introduction}

Matrices are fundamental to modern artificial intelligence. Data are stored in matrices and
tensors, and the principal operations in neural networks are matrix transformations. Most
learned weight matrices, however, are treated as arbitrary finite arrays. This can require many
parameters and can obscure the relationships discovered by a model.

Riordan arrays suggest a different possibility. A Riordan array is not merely an infinite
matrix. It is an infinite matrix whose entries are governed by two formal power series and whose
multiplication is described by multiplication and composition of those
series~\cite{shapiro1991,sprugnoli1994}. It therefore gives a compact algebraic language for
transformations that can be applied at every finite data length.

The guiding question is:
\begin{quote}
\emph{Can Riordan arrays provide useful, learnable, and interpretable structured operators for
sequential, hierarchical, or recursively generated data?}
\end{quote}

There is strong precedent for asking this question. Convolutional layers use Toeplitz-type
structure~\cite{bai2018}; structured state-space models use special matrix parameterizations to
process long sequences~\cite{gu2020hippo,gu2022s4,smith2023}; and neural operators seek models
that act consistently across different discretizations~\cite{kovachki2023}. Recent work has also
shown that structured matrices can replace dense layers while reducing parameters or
computation~\cite{dao2022,liu2022,qiu2024}. Riordan arrays offer a structure that is distinct
from these examples: their fundamental operation is $X(z)\mapsto g(z)X(f(z))$.

\paragraph{Contributions.}
This note is a research proposal with preliminary observations, not a report of experiments. It
makes four contributions.
\begin{enumerate}
\item It collects, in one place and in notation suited to data analysis, the definitions needed
to state the proposal: Riordan arrays, the Riordan group and its standard subgroups, $A$- and
$Z$-sequences, production matrices, and pseudo-involutions (Section~\ref{sec:prelim}).
\item It clarifies what an infinite Riordan array means for finite data, and what the
substitution $X\mapsto X(f)$ does and does not do to a sequence
(Sections~\ref{sec:finite} and~\ref{sec:why}).
\item It records four elementary propositions on identifiability, sums, invertibility and
conditioning, and computational cost (Section~\ref{sec:results}). These settle the easy parts of
several questions and isolate the parts that remain open.
\item It proposes candidate architectures, application domains, open questions, and a
falsifiable experimental program (Sections~\ref{sec:arch}--\ref{sec:exp}).
\end{enumerate}

\paragraph{Notation.}
We write $x=(x_0,x_1,\dots)$ for a data sequence and $X(z)$ for its generating function, and
$y$, $Y(z)$ for the output. The letters $A$ and $Z$ are reserved for the $A$-sequence and
$Z$-sequence of a Riordan array. $R_N$ denotes the leading $N\times N$ truncation of an
array~$R$.

\section{Preliminaries}\label{sec:prelim}

Throughout, $\K$ is a field of characteristic zero, usually $\R$ or $\C$, and $\K[[z]]$ is the
ring of formal power series over $\K$. We write $[z^n]F(z)$ for the coefficient of $z^n$ in
$F(z)$. All series are formal; convergence matters only when numerical stability is discussed.

\subsection{Riordan arrays and the fundamental theorem}

\begin{definition}\label{def:riordan}
A \emph{proper ordinary Riordan array} is a pair $R=(g(z),f(z))$ of formal power series with
$g(0)\neq 0$, $f(0)=0$, and $f'(0)\neq 0$. It is identified with the infinite matrix whose
entries are
\[
r_{n,k} = [z^n]\, g(z)\, f(z)^k, \qquad n,k\ge 0 .
\]
\end{definition}

Thus column $k$ has generating function $g(z)f(z)^k$. The condition $f(0)=0$ makes $R$ lower
triangular. The diagonal entries are $g(0)f'(0)^k$, so the other two conditions make $R$
invertible.

\begin{theorem}[Fundamental theorem of Riordan arrays~\cite{shapiro1991,sprugnoli1994}]\label{thm:ftra}
Let $x=(x_0,x_1,\dots)^{T}$ have ordinary generating function $X(z)$. If $y=Rx$, then
\begin{equation}\label{eq:ftra}
Y(z) = \sum_{n\ge 0} y_n z^n = g(z)\,X(f(z)).
\end{equation}
\end{theorem}

\begin{proof}
Summing columns, $Y(z)=\sum_{k\ge 0} x_k\, g(z)f(z)^k = g(z)X(f(z))$.
\end{proof}

\subsection{The Riordan group}\label{sec:group}

Matrix multiplication of Riordan arrays corresponds to the product
\[
(g,f)\,(h,\ell) = \bigl( g(z)\,h(f(z)),\; \ell(f(z)) \bigr).
\]
The identity is $(1,z)$, and
\[
(g,f)^{-1} = \left( \frac{1}{g(\fbar(z))},\; \fbar(z) \right),
\]
where $\fbar$ is the \emph{compositional inverse} of $f$, the unique series with
$f(\fbar(z))=\fbar(f(z))=z$. It exists because $f(0)=0$ and $f'(0)\neq 0$. The proper Riordan
arrays therefore form a group, the \emph{Riordan group}~\cite{shapiro1991}.

\subsection{Standard subgroups}\label{sec:subgroups}

The following families are subgroups of the Riordan group, and each imposes a different
hypothesis on a data transformation.

\begin{center}
\begin{tabular}{@{}lll@{}}
\toprule
Name & Form & Effect on a sequence\\
\midrule
Appell & $(g(z),\,z)$ & Multiplication by $g$: a causal convolution\\
Lagrange (associated) & $(1,\,f(z))$ & Pure substitution $X(z)\mapsto X(f(z))$\\
Bell (renewal)~\cite{rogers1978} & $(g(z),\,zg(z))$ & One series controls both filter and substitution\\
Derivative & $(f'(z),\,f(z))$ & Filter tied to the derivative of the substitution\\
Hitting-time & $(zf'(z)/f(z),\,f(z))$ & Filter tied to the logarithmic derivative of $f$\\
Checkerboard & $g$ even, $f$ odd & Entries vanish when $n-k$ is odd\\
\bottomrule
\end{tabular}
\end{center}

Every Riordan array factors as $(g,f)=(g,z)(1,f)$, an Appell array times a Lagrange array.

\subsection{The \texorpdfstring{$A$}{A}-sequence and \texorpdfstring{$Z$}{Z}-sequence}\label{sec:AZ}

Riordan arrays can be generated row by row, without reference to $g$ and $f$.

\begin{theorem}[Rogers~\cite{rogers1978}; Merlini, Rogers, Sprugnoli and Verri~\cite{merlini1997}; He and Sprugnoli~\cite{he2009}]\label{thm:AZ}
A lower-triangular array $(r_{n,k})$ with $r_{0,0}\neq 0$ is a proper Riordan array if and only
if there exist sequences $A=(a_0,a_1,a_2,\dots)$ with $a_0\neq 0$ and $Z=(z_0,z_1,z_2,\dots)$
such that, for all $n,k\ge 0$,
\[
r_{n+1,\,k+1} = \sum_{j\ge 0} a_j\, r_{n,\,k+j},
\qquad
r_{n+1,\,0} = \sum_{j\ge 0} z_j\, r_{n,\,j}.
\]
\end{theorem}

The sequence $A$ is the \emph{$A$-sequence} and $Z$ is the \emph{$Z$-sequence} of the array.
Writing $A(t)$ and $Z(t)$ for their generating functions, they are related to $(g,f)$ by
\[
f(z) = z\,A(f(z)), \qquad g(z) = \frac{g(0)}{1 - z\,Z(f(z))},
\qquad\text{equivalently}\qquad A(t) = \frac{t}{\fbar(t)} .
\]
The $A$-sequence says how each entry outside column $0$ is built from the entries above and to
its left; the $Z$-sequence says how column $0$ evolves. For the Pascal array, $A(t)=1+t$ and
$Z(t)=1$, which is Pascal's rule.

We call an $A$- or $Z$-sequence \emph{short} if it has finitely many nonzero terms, and
\emph{sparse} if most of its leading terms vanish. If both sequences have at most $s$ nonzero
terms, row $n+1$ is obtained from row $n$ in $O(ns)$ operations.

\subsection{Production matrices}\label{sec:production}

\begin{definition}[\cite{deutsch2005,deutsch2009}]
Let $\overline{R}$ denote $R$ with its first row deleted. The \emph{production matrix} of $R$ is
$P=R^{-1}\overline{R}$. Equivalently $\overline{R}=RP$: row $n+1$ of $R$ equals row $n$ of $R$
multiplied by $P$.
\end{definition}

For a Riordan array, Theorem~\ref{thm:AZ} says exactly that $P$ has the $Z$-sequence as its
first column and shifted copies of the $A$-sequence in the remaining columns:
\[
P = \begin{pmatrix}
z_0 & a_0 & 0 & 0 & \cdots \\
z_1 & a_1 & a_0 & 0 & \cdots \\
z_2 & a_2 & a_1 & a_0 & \cdots \\
z_3 & a_3 & a_2 & a_1 & \cdots \\
\vdots & \vdots & \vdots & \vdots & \ddots
\end{pmatrix}.
\]
Conversely, any matrix of this shape with $a_0\neq 0$, together with a value $r_{0,0}\neq 0$,
generates a unique Riordan array. The production matrix is tridiagonal exactly when $a_j=0$ for
$j\ge 3$ and $z_j=0$ for $j\ge 2$; this is the case linked to three-term recurrences and
orthogonal polynomials.

\subsection{Pseudo-involutions}\label{sec:pseudo}

Let $M=(1,-z)$, the diagonal matrix $\operatorname{diag}(1,-1,1,-1,\dots)$, so that $M^2=I$. A
Riordan array $R$ is an \emph{involution} if $R^2=I$, and a \emph{pseudo-involution} if $RM$ is
an involution~\cite{cameron2005}. Since $(RM)^2=I$ is equivalent to $R^{-1}=MRM$, a
pseudo-involution satisfies
\[
\bigl(R^{-1}\bigr)_{n,k} = (-1)^{\,n-k}\, r_{n,k}.
\]
In terms of the defining series, $R=(g,f)$ is a pseudo-involution if and only if
$g(z)\,g(-f(z))=1$ and $f(-f(z))=-z$~\cite{cameron2005}. The Pascal array is the standard
example: its inverse has entries $(-1)^{n-k}\binom{n}{k}$.

\subsection{Exponential Riordan arrays}\label{sec:exponential}

An \emph{exponential Riordan array} $[g(z),f(z)]$ is defined by the same pair of series but with
entries
\[
r_{n,k} = \frac{n!}{k!}\,[z^n]\, g(z)\, f(z)^k .
\]
It acts on exponential generating functions $\widehat X(z)=\sum_{n\ge 0} x_n z^n/n!$ by
$\widehat Y(z)=g(z)\widehat X(f(z))$, and the group law is unchanged. In this setting the Pascal
array is $[e^z,z]$.

\section{Riordan operators on finite data}\label{sec:finite}

By Theorem~\ref{thm:ftra}, a Riordan array acts on a data sequence by
$X(z)\mapsto g(z)X(f(z))$, or coefficientwise by
\[
y_n = \sum_{k=0}^{n} r_{n,k}\, x_k .
\]
The output at level $n$ depends only on inputs at levels $0,1,\dots,n$. This is the type of
causal or graded dependence that occurs in time series, event histories, rooted trees, and many
recursive processes. By Section~\ref{sec:group}, both the composition and the inversion of these
transformations are explicit.

\subsection{What ``infinite data'' should mean}

An infinite Riordan array does not physically store an infinite amount of arbitrary data. If $g$
and $f$ are arbitrary series, they themselves may require infinitely many coefficients. The
practical advantage appears when $g$ and $f$ have compact descriptions, such as rational or
algebraic forms, short recurrences, short or sparse $A$- and $Z$-sequences, or finite-dimensional
parameterizations.

The more defensible interpretation is that a Riordan array defines an unbounded,
\emph{length-consistent} operator. For every $N$, let $\pi_N(R)=R_N$ denote the leading
$N\times N$ truncation.

\begin{lemma}[Length consistency]\label{lem:length}
Let $R$ and $S$ be infinite lower-triangular matrices and $x$ a sequence. Then:
\begin{enumerate}
\item $\pi_N(RS)=\pi_N(R)\,\pi_N(S)$, and $\pi_N(R^{-1})=\pi_N(R)^{-1}$ when $R$ is invertible;
\item the first $N$ entries of $Rx$ equal $R_N$ applied to the first $N$ entries of $x$;
\item if $R=(g,f)$ is a Riordan array, $R_N$ depends only on $g \bmod z^N$ and $f \bmod z^N$.
\end{enumerate}
\end{lemma}

\begin{proof}
Parts 1 and 2 hold because entry $(n,k)$ of a product of lower-triangular matrices involves only
indices between $k$ and $n$. Part 3 follows from Definition~\ref{def:riordan}.
\end{proof}

The same algebraic rule can therefore act on sequences of length 20, 200, or 2000, and extending
a sequence never changes the outputs already computed. All actual computations remain finite,
but they are restrictions of a common formal operator. This viewpoint is related in spirit to
neural operators, which learn mappings between function spaces and apply common parameters
across discretizations~\cite{kovachki2023}.

\section{Why the Riordan structure may be useful}\label{sec:why}

\subsection{Convolution as a special case}

For an Appell array $(g(z),z)$, equation~\eqref{eq:ftra} becomes $Y(z)=g(z)X(z)$. Writing
$g(z)=\sum_{j\ge 0} g_j z^j$ gives
\[
y_n = \sum_{k=0}^{n} g_{n-k}\, x_k .
\]
This is a causal convolution. Hence Riordan transformations contain ordinary one-dimensional
causal convolutional filters, and a general Riordan array may be regarded as a structured
extension of a causal convolutional layer.

\subsection{What the substitution adds}\label{sec:substitution}

Write $f(z)=z\,h(z)$ with $h(0)=f'(0)\neq 0$. Column $k$ of $R$ has generating function
$z^k g(z)h(z)^k$, so
\[
y_n = \sum_{k=0}^{n} \kappa^{(k)}_{\,n-k}\; x_k ,
\qquad
\kappa^{(k)}(z) = g(z)\, h(z)^k .
\]
A Riordan operator is therefore a causal convolution whose kernel depends on the index $k$ of
the input it is applied to, with the kernels constrained to be $g$ times the $k$-th power of a
single series $h$. Three cases illustrate the range.
\begin{itemize}
\item \textbf{Appell, $h=1$.} The kernel is the same for every $k$; this is ordinary
convolution.
\item \textbf{Linear substitution, $f(z)=cz$.} Then $y_n=\sum_{k} g_{n-k}\,c^k x_k$: a geometric
reweighting of the input, which is exponential discounting when $|c|<1$, followed by
convolution.
\item \textbf{Pascal, $g=h=1/(1-z)$.} The kernel applied to $x_k$ is $(1-z)^{-(k+1)}$, so later
inputs are spread by progressively heavier kernels.
\end{itemize}

It is equally important to say what the substitution does not do. Because $f'(0)\neq 0$, it
cannot dilate or subsample the index; a map such as $X(z)\mapsto X(z^2)$ is excluded. Describing
$X\mapsto X(f)$ as a change of time scale would be misleading. The accurate description is
index-dependent filtering governed by one series.

\subsection{A familiar example}

The Pascal array is
\[
P = \left( \frac{1}{1-z},\; \frac{z}{1-z} \right),
\qquad
p_{n,k} = \binom{n}{k}.
\]
It performs the binomial transform $y_n=\sum_{k=0}^{n}\binom{n}{k}x_k$. An AI model could use
this transformation as a fixed feature map. More interestingly, it could learn a deformation
$(g_\theta(z),f_\theta(z))$ from data. The learned pair would provide an algebraic description
of how the model weights earlier observations and propagates them across levels.

\subsection{Parameter efficiency}

An arbitrary $N\times N$ lower-triangular matrix has $N(N+1)/2$ independent entries. By
Lemma~\ref{lem:length}, the truncation $R_N$ of a Riordan array is determined by
$g_0,\dots,g_{N-1}$ and $f_1,\dots,f_{N-1}$, that is, by $2N-1$ numbers. A rational or
low-degree parameterization may use still fewer.

This restriction is both an advantage and a risk. It can improve statistical efficiency when the
data possess the corresponding structure, but it can cause underfitting when they do not. In
machine-learning language, the Riordan condition supplies an inductive bias; its value must be
tested rather than assumed.

\subsection{Interpretability}

An unrestricted learned matrix is usually difficult to interpret. A learned Riordan
transformation can instead be examined through:
\begin{itemize}
\item $g(z)$, which governs weighting and filtering and is the generating function of the first
column;
\item $f(z)$, which governs how the kernel changes with the input index
(Section~\ref{sec:substitution});
\item the $A$-sequence, which describes the recursive generation of columns
(Section~\ref{sec:AZ});
\item the $Z$-sequence, which describes the evolution of the first column; and
\item the production matrix, which describes the transition from one row or level to the next
(Section~\ref{sec:production}).
\end{itemize}
These objects may reveal the recurrence or propagation rule learned from the data. This creates
the possibility of combining prediction with symbolic interpretation.

\section{Four elementary observations}\label{sec:results}

The propositions in this section are immediate from the definitions. We record them because each
one settles the easy part of a question raised later and shows where the real difficulty lies.

\subsection{Identifiability}

\begin{proposition}\label{prop:ident}
Let $R=(g,f)$ be a proper Riordan array.
\begin{enumerate}
\item The response of $R$ to the input $X(z)=1$ is $g(z)$, and its response to $X(z)=z$ is
$g(z)f(z)$. Hence $g$ and $f=(gf)/g$ are determined by two input--output pairs.
\item The truncation $R_N$ determines $g \bmod z^N$ and $f \bmod z^N$, and conversely.
\item A single input--output pair does not determine $(g,f)$ in general.
\end{enumerate}
\end{proposition}

\begin{proof}
Part 1 is equation~\eqref{eq:ftra}. Part 2 follows from part 1 and Lemma~\ref{lem:length}, since
$g$ is invertible modulo $z^N$. For part 3, the input $X=1$ carries no information about $f$.
\end{proof}

\begin{remark}[Where non-uniqueness actually arises]\label{rem:gauge}
A single Riordan operator has no rescaling ambiguity: the map $(g,f)\mapsto R$ is injective.
Ambiguity appears only in compositions. For any diagonal Riordan array $D=(c,dz)$ with $c,d>0$,
and any positively homogeneous coordinatewise activation $\sigma$ such as ReLU,
\[
R_2\, \sigma(R_1 x) = \bigl(R_2 D^{-1}\bigr)\, \sigma\bigl(D R_1 x\bigr).
\]
Normalizing $g(0)=1$ and $f'(0)=1$ in every layer but the last removes this freedom. What
Proposition~\ref{prop:ident} leaves open is \emph{stability}: recovering $f$ requires dividing
by $g$, and it is not clear how noise in the observations propagates to high-order coefficients.
\end{remark}

\subsection{Sums leave the Riordan class}

\begin{proposition}\label{prop:sums}
The set of Riordan arrays is not closed under addition.
\end{proposition}

\begin{proof}
Let $S=(1,z)+(1,2z)$, the diagonal matrix with entries $1+2^k$. Its first three columns have
generating functions $2$, $3z$, and $5z^2$. If $S$ were $(g,f)$, then $g=2$ and $f=3z/2$, which
forces the third column to be $9z^2/2$.
\end{proof}

Products of Riordan operators collapse to a single Riordan operator by the group law, but sums
do not. A sum of $r$ Riordan operators has about $r(2N-1)$ parameters and forms a strictly
larger class, in the way that several convolutional filters are more expressive than one.

\subsection{Determinant, inverse, and conditioning}\label{sec:conditioning}

\begin{proposition}\label{prop:det}
Let $R=(g,f)$ be a proper Riordan array over $\R$ or $\C$, and let $M_N=\pi_N(M)$.
\begin{enumerate}
\item $\det R_N = g(0)^N f'(0)^{N(N-1)/2}$.
\item $(R_N)^{-1}$ is the $N\times N$ truncation of the explicit inverse in
Section~\ref{sec:group}.
\item If $R$ is a pseudo-involution, then $(R_N)^{-1}=M_N R_N M_N$. Consequently
$\|(R_N)^{-1}\|_2=\|R_N\|_2$, and the spectral condition number is
$\kappa_2(R_N)=\|R_N\|_2^2$.
\end{enumerate}
\end{proposition}

\begin{proof}
Part 1 holds because $R_N$ is triangular with diagonal entries $g(0)f'(0)^k$. Part 2 is
Lemma~\ref{lem:length}. For part 3, truncate $R^{-1}=MRM$ and note that $M_N$ is diagonal and
orthogonal.
\end{proof}

Part 1 gives the log-determinant of a Riordan layer in closed form, which is the quantity needed
by normalizing flows~\cite{papamakarios2021}; with $g(0)=f'(0)=1$ the layer is volume
preserving. Part 3 is a warning as much as a convenience. A pseudo-involutory layer has a free
inverse, but it is well conditioned only if its norm is small. For the Pascal array the largest
entry of $P_N$ is a central binomial coefficient, so $\kappa_2(P_N)$ grows at least like
$4^N/N$.

\subsection{Applying a Riordan layer without forming its matrix}

Let $\Mult(N)$ denote the cost of multiplying two polynomials of degree less than $N$; with the
fast Fourier transform, $\Mult(N)=O(N\log N)$.

\begin{proposition}\label{prop:cost}
Given $g$ and $f$ modulo $z^N$ and a vector $x$ of length $N$, the product $R_N x$ is the
coefficient vector of $g(z)X(f(z)) \bmod z^N$. It can be computed with
\begin{enumerate}
\item $\Mult(N)$ operations if $R$ is Appell;
\item $O\bigl(\Mult(N)\,(N\log N)^{1/2}\bigr)$ operations in general, by the algorithm of Brent
and Kung~\cite{brent1978};
\item $O\bigl(\Mult(N)\log N\bigr)$ operations in general, by the algorithm of Kinoshita and
Li~\cite{kinoshita2024}.
\end{enumerate}
The explicit matrix $R_N$ can be formed in $O(N^2 s)$ operations from $A$- and $Z$-sequences
with at most $s$ nonzero terms.
\end{proposition}

\begin{proof}
The first statement is Theorem~\ref{thm:ftra} with Lemma~\ref{lem:length}. The costs are those
of one truncated power-series multiplication and one truncated composition. The last statement
follows from the recurrences in Theorem~\ref{thm:AZ}.
\end{proof}

The asymptotic question of whether a Riordan layer can be applied in quasi-linear time is
therefore already answered. The open question is practical: whether these algebraic algorithms
are numerically stable in floating point, differentiable in a form that automatic-differentiation
frameworks handle well, and competitive with a dense $N\times N$ product on current hardware at
realistic $N$.

\section{Candidate AI architectures}\label{sec:arch}

\subsection{A learnable Riordan layer}\label{sec:layer}

The most direct construction replaces a dense linear layer by
\begin{equation}\label{eq:layer}
y = \sigma\bigl( R_N(g_\theta, f_\theta)\, x + b \bigr),
\end{equation}
where $\sigma$ is an activation function, $b$ is a bias vector, and $\theta$ is learned by
gradient descent. A convenient parameterization is
\[
g_\theta(z) = s_g \exp\Bigl( \sum_{j=1}^{p} \alpha_j z^j \Bigr),
\qquad
f_\theta(z) = s_f\, z \exp\Bigl( \sum_{j=1}^{q} \beta_j z^j \Bigr).
\]
The pair is proper exactly when $s_g s_f\neq 0$. To guarantee this throughout training, one can
fix $s_g=s_f=1$, which is the normalization of Remark~\ref{rem:gauge}, or write each scale as
$\pm\exp(\tau)$ with a learned $\tau$. Every series with nonzero constant term has this
exponential form, so $p=q=N-1$ reparameterizes all of $R_N$, and smaller $p$ and $q$ give
low-dimensional subfamilies. Alternative parameterizations could use rational functions,
continued fractions, orthogonal-polynomial bases, or directly learned $A$- and $Z$-sequences.

Without an intervening nonlinearity, a product of Riordan layers collapses to one Riordan layer
by the group law. This gives exact simplification, but it also means that depth alone does not
add expressivity to a purely linear Riordan network. Nonlinear activations, channel mixing,
residual connections, sums of Riordan operators (Proposition~\ref{prop:sums}), or
input-dependent Riordan parameters would be needed for a genuinely deep architecture.

\subsection{A production-matrix recurrent model}\label{sec:recurrent}

Instead of constructing $R_N$ directly, one could learn short $A$- and $Z$-sequences, which
define a production matrix $P$ (Section~\ref{sec:production}). The model then generates the
operator one row at a time. Writing $\rho_n$ for row $n$ of $R$,
\[
\rho_{n+1} = \rho_n P,
\qquad
y_n = \rho_n \cdot (x_0, \dots, x_n)^{T}.
\]
With sequences of at most $s$ nonzero terms, each step costs $O(ns)$ operations. This direction
may be particularly natural for streaming data.

One distinction from structured state-space
models~\cite{gu2020hippo,gu2022s4,smith2023} should be stated plainly. In those models a
fixed-size state summarizes the data seen so far. Here the recurrent state $\rho_n$ is a row of
\emph{weights}: it evolves independently of the input, and the model must retain the input
history. A fixed-size data state of dimension $d$ exists only if every block of $R$ lying below
and to the left of the diagonal has rank at most $d$. This holds for Appell arrays with rational
$g$, which are classical recursive filters. It fails for the Pascal array, whose response to an
impulse at position $k$ is a polynomial of degree $k$ in $n$. Which Riordan operators admit
bounded-memory realizations is an open question (Section~\ref{sec:questions}).

Two versions should be investigated:
\begin{enumerate}
\item learn a short $A$-sequence and $Z$-sequence and derive the associated array; or
\item learn a sparse lower-Hessenberg matrix $P$ and penalize its deviation from the Riordan
shape of Section~\ref{sec:production}, for example by the variance of its entries along each
diagonal outside the first column.
\end{enumerate}
The second version would permit the data to determine whether exact Riordan structure is too
restrictive.

\subsection{Multichannel and block-Riordan models}

Real observations usually contain several features at each time or level. Let
$X(z)=\sum_{n\ge 0} x_n z^n$ with $x_n\in\R^d$. A simple model applies a common scalar
substitution and a matrix-valued filter:
\[
Y(z) = G(z)\, X(f(z)),
\]
where $G(z)$ is a $d\times d$ matrix of formal series with $G(0)$ invertible. Because $f$ is
scalar, the product rule of Section~\ref{sec:group} still holds with $G$ in place of $g$. A less
expensive alternative applies scalar Riordan operators to individual channels and follows them
with a learned finite-dimensional channel-mixing matrix.

Multivariate Riordan arrays may also be relevant when data have more than one meaningful
grading, such as time and location, or degree and depth in a network. The increased expressivity
must be balanced against computational cost and interpretability.

\subsection{Input-dependent Riordan layers}

In an ordinary Riordan layer, the same pair $(g,f)$ is used for every input. A more flexible
model could let a small conditioning network produce a pair $(g_{\theta,x},f_{\theta,x})$ that
depends on the input $x$. The operator would then adapt to the observed sequence, in analogy
with attention~\cite{vaswani2017} or selective state-space mechanisms~\cite{gu2023mamba}. The
conditions $g(0)\neq 0$ and $f'(0)\neq 0$ can be enforced by the parameterization of
Section~\ref{sec:layer}.

This construction is no longer a single fixed linear operator, but each instance remains
algebraically interpretable. It may offer a useful compromise between rigid Riordan structure
and unrestricted attention.

\subsection{Invertible layers and pseudo-involutions}

The explicit inverse of a Riordan array suggests applications to reversible
networks~\cite{gomez2017}, normalizing flows and density
estimation~\cite{dinh2014,papamakarios2021}, encoder--decoder systems, and data reconstruction.
Proposition~\ref{prop:det} supplies the two ingredients such models need: an exact inverse and a
closed-form log-determinant. If coefficient growth is controlled, the same learned algebraic
object can serve as an encoder while its group inverse serves as the decoder.

Pseudo-involutions (Section~\ref{sec:pseudo}) provide an especially interesting constraint,
because the inverse is obtained from the forward operator by a fixed sign conjugation,
$R^{-1}=MRM$. A pseudo-involutory layer could support:
\begin{itemize}
\item a tied encoder and decoder;
\item exact cycle-consistency constraints;
\item reversible transformations with fewer independently learned parameters; and
\item paired or contrastive transformations generated by $R$ and $R^{-1}$.
\end{itemize}

The constraint can be imposed without a penalty term. For any Riordan array $R$, the matrix
$RMR^{-1}$ is an involution, so $Q=RMR^{-1}M$ is a pseudo-involution. Learning an unconstrained
$R$ and using $Q$ as the layer therefore stays inside the pseudo-involutions throughout
training. Whether every pseudo-involution has this form is a question raised
in~\cite{cameron2005}. By Proposition~\ref{prop:det}, the conditioning of such a layer is the
square of its norm, so norm control is essential.

This direction connects the proposed AI work directly with current research on factorization,
products, and subgroups of pseudo-involutions.

\subsection{Riordan-equivariant models and data augmentation}\label{sec:equivariant}

If a collection of Riordan transformations represents meaningful symmetries of a
data-generating process, one could require a model $F$ to satisfy $F(Rx)=R'F(x)$ for suitable
input and output actions, as in group-equivariant networks~\cite{cohen2016}. Alternatively,
elements from a selected Riordan subgroup could generate augmented training examples.

This proposal must be used carefully: an arbitrary Riordan transformation need not preserve the
label or scientific meaning of an observation. The subgroup must arise from a defensible
invariance of the application. The Appell, Bell, Lagrange, checkerboard, and hitting-time
subgroups of Section~\ref{sec:subgroups}, or a family of pseudo-involutions, would impose
different hypotheses about the data.

\section{Possible application domains}\label{sec:applications}

\subsection{Time series and event histories}

The lower-triangular action is naturally causal. Possible tasks include forecasting, smoothing,
denoising, change-point detection, and classification of event sequences. Appell arrays provide
convolutional baselines, while a general $f$ tests whether index-dependent filtering
(Section~\ref{sec:substitution}) improves performance.

\subsection{Hierarchical and recursively generated data}

Riordan arrays already encode many families of lattice paths and trees. They may be appropriate
when an observation at level $n$ is produced from distributions at earlier levels. Examples
include rooted-tree statistics, branching processes, genealogical data, and recursively
generated networks. Production matrices may be more natural than direct matrix multiplication in
these settings, because they describe the passage from one level to the next.

\subsection{Moment, count, and probability data}

Exponential Riordan arrays (Section~\ref{sec:exponential}) may be preferable when factorial
normalization is intrinsic, as in labelled structures, moment sequences, event counts, and some
probabilistic models. Barry's study of sigmoid functions in the derivative subgroup of
exponential Riordan arrays gives one existing connection between Riordan theory and functions
used in machine learning~\cite{barry2017}, although it does not develop the learned operator
architectures proposed here.

Orthogonal polynomials associated with production matrices may also provide data-adapted bases.
A model might learn a structured transformation and then analyze whether its production matrix
is tridiagonal or nearly tridiagonal (Section~\ref{sec:production}), linking the learned
representation to moment functionals and orthogonality.

\subsection{Longitudinal social data}

This application is the most speculative of the five. For longitudinal household data, an array
could encode how attributes at one generation or socioeconomic level contribute to later
distributions. If two histories are represented by operators $R_1$ and $R_2$, their product
$R_1R_2$ provides a noncommutative model of sequential combination, and the order of the factors
must be given a substantive interpretation before the model means anything.

Such a study would require careful separation of descriptive modeling from causal claims.
Fairness, identifiability, privacy, and the social meaning of the chosen ordering would be
essential parts of it.

\subsection{Anomaly detection and structure discovery}

Given an observed $N\times N$ lower-triangular dependency matrix $W$, one may seek the closest
Riordan approximation
\[
\min_{g,\,f}\; \bigl\| W - R_N(g,f) \bigr\| .
\]
Proposition~\ref{prop:ident} supplies a natural initial guess: read $g$ from the first column of
$W$ and $f$ from the quotient of the first two columns, then refine. The residual could measure
departure from a simple recursive law. Changes in the fitted $g$, $f$, $A$, or $Z$ parameters
over time could signal regime changes or anomalies. A ``soft Riordan'' penalty could likewise
regularize a dense model while permitting deviations supported by the data.

\section{Open research questions}\label{sec:questions}

The proposed architectures lead to questions that combine Riordan theory, approximation theory,
numerical analysis, and machine learning. Section~\ref{sec:results} answers the easy parts of
several of them; the questions below are what remains.

\begin{question}[Approximation]
Which causal linear operators, or which finite lower-triangular matrices, can be approximated
efficiently by a single Riordan array? How much does the approximation class expand when sums of
$r$ Riordan operators (Proposition~\ref{prop:sums}), or alternating products of Riordan and
channel-mixing operators, are allowed?
\end{question}

\begin{question}[Stable identifiability]
Proposition~\ref{prop:ident} recovers $g$ and $f$ exactly from two noiseless input--output
pairs. How do errors in the observations propagate to the coefficients of $g$ and $f$, as a
function of their order? What sample size and what input distributions give stable recovery?
\end{question}

\begin{question}[Length generalization]
If a pair $(g,f)$ is learned from $N\times N$ truncations, under what assumptions will the
resulting operator make stable predictions at orders greater than $N$?
\end{question}

\begin{question}[Stability]
How do the analytic properties of $g$ and $f$, such as their radii of convergence, control
norms, condition numbers, gradient growth, and error propagation in $R_N(g,f)$ as $N$ increases?
Proposition~\ref{prop:det} reduces this to a norm bound in the pseudo-involutory case.
\end{question}

\begin{question}[Practical fast algorithms]
Proposition~\ref{prop:cost} shows that a Riordan layer can be applied in quasi-linear time in
exact arithmetic. Are the underlying composition algorithms numerically stable in floating point
and efficient on current hardware? At what $N$ do they overtake a dense product, and which
parameterizations of $g$ and $f$ make them simplest?
\end{question}

\begin{question}[Bounded memory]
Which Riordan arrays have the property that every block below and to the left of the diagonal
has rank at most $d$, so that the operator admits a streaming realization with a state of
dimension $d$ (Section~\ref{sec:recurrent})? Can this class be described in terms of $g$ and
$f$, or of the $A$- and $Z$-sequences?
\end{question}

\begin{question}[Subgroup selection]
Do particular data types naturally select the Appell, Lagrange, Bell, checkerboard,
hitting-time, or pseudo-involution families? Can subgroup selection itself be learned or
validated statistically?
\end{question}

\begin{question}[Symbolic discovery]
Can a numerically learned pair $(g,f)$ be converted into a conjectured rational, algebraic, or
differential description? Can a learning system identify a short $A$-sequence, $Z$-sequence, or
production rule and then certify it symbolically?
\end{question}

The last question reverses the direction of earlier work using machine learning to study Riordan
arrays [citation to be supplied]. Instead of asking a learning system to detect properties of
known arrays, one asks a data-driven system to discover a Riordan law and then uses Riordan
theory to interpret or prove the law.

\section{A proposed initial experimental program}\label{sec:exp}

The first study should be deliberately modest. Its purpose would be to determine whether Riordan
structure contributes anything beyond familiar baselines. No experiments have yet been run.

\subsection{Phase I: controlled synthetic experiments}

Generate data from known Riordan operators and add controlled noise. Ask whether the learning
procedure can:
\begin{enumerate}
\item recover $g$ and $f$, or an equivalent $A$- and $Z$-description, more accurately than the
direct read-off of Proposition~\ref{prop:ident};
\item predict unseen output coefficients;
\item distinguish Appell structure from genuinely compositional structure;
\item remain stable when extrapolating beyond the training truncation; and
\item identify when the generating operator is not Riordan.
\end{enumerate}
Synthetic tests should include rational examples such as the Pascal array, algebraic examples
such as the Catalan array $(C(z),zC(z))$ with $C(z)$ the Catalan generating function, examples
with sparse $A$- and $Z$-sequences, pseudo-involutions, and adversarial non-Riordan matrices
such as the sum in Proposition~\ref{prop:sums}.

\subsection{Phase II: benchmark comparisons}

For an ordered-data benchmark, compare:
\begin{enumerate}
\item a dense lower-triangular layer;
\item a causal convolution or Toeplitz layer~\cite{bai2018};
\item a fixed Riordan feature transform;
\item a learnable Riordan layer (Section~\ref{sec:layer});
\item a production-matrix recurrent model (Section~\ref{sec:recurrent});
\item a structured state-space model~\cite{gu2022s4,smith2023}; and
\item a hybrid model combining a Riordan layer with a small attention or channel-mixing
component.
\end{enumerate}
The evaluation should report prediction error, parameter count, wall-clock time, memory use,
numerical conditioning, robustness to noise, and performance on sequences longer than those used
in training. Interpretability should be assessed by examining whether the fitted $g$, $f$, $A$,
and $Z$ objects are stable across repeated training runs.

\subsection{Phase III: ablation and falsification}\label{sec:ablation}

A useful study should attempt to falsify the Riordan hypothesis. Relevant ablations include
fixing $f(z)=z$, replacing composition by a generic lower-triangular matrix, removing the group
constraint, and matching parameter counts across all models. If the general Riordan layer does
not improve on the Appell/Toeplitz case, that result would show that convolution, not
composition, is responsible for the performance.

\section{Computational and conceptual limitations}

Several obstacles should be addressed from the beginning.
\begin{enumerate}
\item \textbf{Not every dataset has a meaningful order.} A Riordan operator is poorly motivated
for tabular observations whose row ordering is arbitrary.
\item \textbf{Coefficient growth can be severe.} Combinatorial arrays can become ill-conditioned
exponentially fast, as the Pascal array does (Section~\ref{sec:conditioning}). Normalization,
rescaling, bounded analytic parameterizations, or exponential versions may be necessary.
\item \textbf{Formal validity is not numerical stability.} The existence of a formal inverse
does not guarantee a well-conditioned inverse at large truncation sizes.
\item \textbf{Truncation limits extrapolation.} A finite training sample cannot identify
arbitrary higher-order coefficients. Reliable extension requires a low-complexity model for $g$
and $f$, together with appropriate regularization.
\item \textbf{Hardware efficiency matters.} Fewer mathematical parameters do not automatically
produce faster GPU computations, and asymptotically fast algorithms
(Proposition~\ref{prop:cost}) may not be the fastest at practical sizes.
\item \textbf{Memory is not automatically bounded.} A general Riordan operator needs the whole
input history (Section~\ref{sec:recurrent}), unlike a state-space model.
\item \textbf{Linear Riordan structure is restrictive.} Applications will likely require
nonlinearities, multiple channels, sums of operators, or hybrid architectures.
\item \textbf{Interpretation is application-dependent.} Although $g$ and $f$ are algebraically
interpretable, their scientific meanings must be established in each domain.
\end{enumerate}

\section{Relationship to existing work}

\paragraph{Riordan arrays.}
The Riordan group was introduced by Shapiro, Getu, Woan, and Woodson~\cite{shapiro1991},
building on the renewal arrays of Rogers~\cite{rogers1978}, and developed as a tool for
combinatorial sums by Sprugnoli~\cite{sprugnoli1994}. The $A$- and $Z$-sequence characterization
is due to Rogers~\cite{rogers1978}, Merlini, Rogers, Sprugnoli, and Verri~\cite{merlini1997},
and He and Sprugnoli~\cite{he2009}. Production matrices were introduced by Deutsch, Ferrari, and
Rinaldi~\cite{deutsch2005,deutsch2009}. Pseudo-involutions were studied by Cameron and
Nkwanta~\cite{cameron2005}, among others. Barry connected exponential Riordan arrays with
sigmoid functions~\cite{barry2017}.

\paragraph{Structured matrices in machine learning.}
Toeplitz, low-rank, sparse, Fourier, and Monarch-type parameterizations have been used to reduce
computation or encode useful structure~\cite{dao2022,liu2022,qiu2024}. Causal convolutional
networks are a strong baseline for sequence modeling~\cite{bai2018}. These results do not
establish that Riordan arrays will be effective, but they establish that the broader strategy is
mathematically and computationally credible.

\paragraph{State-space and operator models.}
HiPPO derives recurrent memory updates from projections onto orthogonal polynomial
bases~\cite{gu2020hippo}, and the structured state-space models S4~\cite{gu2022s4},
S5~\cite{smith2023}, and Mamba~\cite{gu2023mamba} build on it. These are the closest relatives
of the production-matrix model of Section~\ref{sec:recurrent}, with the difference in state
noted there. Neural operators provide a separate but related framework for learning
transformations not tied to a single finite discretization~\cite{kovachki2023}.

\paragraph{Invertible and equivariant models.}
Reversible networks~\cite{gomez2017} and normalizing flows~\cite{dinh2014,papamakarios2021} rely
on layers with tractable inverses and Jacobian determinants, often triangular ones;
Proposition~\ref{prop:det} places Riordan layers in that family. Group-equivariant
networks~\cite{cohen2016} are the model for Section~\ref{sec:equivariant}.

\paragraph{Computer algebra.}
Fast composition of power series, the computational core of a Riordan layer, goes back to Brent
and Kung~\cite{brent1978} and was recently brought to quasi-linear time by Kinoshita and
Li~\cite{kinoshita2024}.

\paragraph{Novelty.}
We searched for prior work on learnable Riordan layers of the form~\eqref{eq:layer}, Riordan
production-matrix recurrent networks, and pseudo-involutory encoder--decoder models, and found
none. The search was not exhaustive, and we make no formal claim of priority. The potentially
original contribution is not the use of an infinite matrix. It is the combination of four
features in a trainable model:
\begin{enumerate}
\item generating-function representation of data;
\item the transformation $X(z)\mapsto g(z)X(f(z))$;
\item Riordan group composition and inversion; and
\item interpretable recurrence data through $A$-sequences, $Z$-sequences, and production
matrices.
\end{enumerate}

\section{Central hypothesis}

The program can be organized around one testable hypothesis:
\begin{quote}
\emph{For data with a genuine causal, graded, or recursive organization, a learned Riordan
operator can provide a more parameter-efficient and interpretable model than an unrestricted
lower-triangular transformation, while retaining accuracy and supporting generalization across
sequence lengths.}
\end{quote}
The hypothesis is falsified, for a given class of data, if the learnable Riordan layer fails to
match the dense baseline at equal accuracy with fewer parameters, or fails to improve on the
Appell case (Section~\ref{sec:ablation}).

\section{Conclusion}

Riordan arrays should not be viewed as containers that hold infinite observational data. Their
real promise is more subtle and more useful: they define structured transformations that extend
coherently to arbitrary finite orders. They combine causal triangular matrices, convolution,
formal substitution, explicit group operations, and recursive descriptions.

This makes them plausible candidates for structured AI models, particularly when the data have
an ordered or recursively generated character. The most compelling first directions are a
learnable Riordan layer, a production-matrix recurrent model driven by $A$- and $Z$-sequences,
and an invertible architecture based on pseudo-involutions. Whether these models improve
prediction must be decided by careful comparison with dense, convolutional, and state-space
baselines. Even if the empirical gains are limited, the project may yield a useful mathematical
theory of approximation, stability, identifiability, and symbolic discovery for
Riordan-structured operators.


\end{document}